\documentclass[10pt,twocolumn,letterpaper]{article}

\usepackage[pagenumbers]{cvpr} % To force page numbers, e.g. for an arXiv version

\definecolor{cvprblue}{rgb}{0.21,0.49,0.74}
\usepackage[pagebackref=false,breaklinks,colorlinks,allcolors=cvprblue]{hyperref}

\def\paperID{5233} % *** Enter the Paper ID here
\def\confName{CVPR}
\def\confYear{2026}

\usepackage{url}
\usepackage{lscape}
\usepackage{array,booktabs,makecell,adjustbox,colortbl}
\newcolumntype{L}[1]{>{\raggedright\arraybackslash}p{#1}} % wrapable left column
\definecolor{mypurple}{RGB}{200,180,250}
\usepackage{wrapfig}
\usepackage{caption}
\usepackage{tcolorbox}
\usepackage{algorithm}
\usepackage{algpseudocode}
\usepackage{amsmath}
\usepackage{graphicx}
\usepackage{xcolor}
\usepackage{wrapfig}
\usepackage{amssymb}    
\tcbuselibrary{listings}
\usepackage{graphicx}

\usepackage{amsmath,amsfonts,bm}

\def\eqref#1{equation~\ref{#1}}
\def\1{\bm{1}}

\DeclareMathAlphabet{\mathsfit}{\encodingdefault}{\sfdefault}{m}{sl}
\SetMathAlphabet{\mathsfit}{bold}{\encodingdefault}{\sfdefault}{bx}{n}

\usepackage{wrapfig}
\usepackage{url}
\tcbuselibrary{breakable}
\usepackage{booktabs}
\usepackage{multirow} 
\usepackage{booktabs}
\usepackage{caption}
\usepackage{subcaption}
\usepackage{booktabs}
\usepackage{multirow}
\usepackage{colortbl}
\usepackage{array}
\usepackage{xcolor}

\newcommand{\ours}{PSO}

\title{Knowing the Answer Isn’t Enough: Fixing Reasoning Path Failures in LVLMs}
\author{
Chaoyang Wang\textsuperscript{1,\thanks{Equal contribution.}}\;\;\;
Yangfan He\textsuperscript{1,\footnotemark[1]}\;\;\;
Yiyang Zhou\textsuperscript{1}\;\;\;
Yixuan Wang\textsuperscript{1}\;\;\; \\
Jiaqi Liu\textsuperscript{1}\;\;\;
Peng Xia\textsuperscript{1}\;\;\;
Zhengzhong Tu\textsuperscript{2}\;\;\;
Mohit Bansal\textsuperscript{1}\;\;\;
Huaxiu Yao\textsuperscript{1} \vspace{2mm}\\
\textsuperscript{1}UNC-Chapel Hill\;\;\;
\textsuperscript{2}Texas A\&M University \\
}

\begin{document}
\maketitle
\begin{abstract}
    We reveal a critical yet underexplored flaw in Large Vision-Language Models (LVLMs): even when these models know the correct answer, they frequently arrive there through incorrect reasoning paths. The core issue is not a lack of knowledge, but a path selection bias within the vast reasoning search space. Although LVLMs are often capable of sampling correct solution trajectories, they disproportionately favor unstable or logically inconsistent ones, leading to erratic and unreliable outcomes. The substantial disparity between Pass@K (with large K) and Pass@1 across numerous models provides compelling evidence that such failures primarily stem from misreasoning rather than ignorance. To systematically investigate and address this issue, we propose \textbf{\ours}\ (Path-Select Optimization), a two-stage post-training framework designed to enhance both the reasoning performance and stability of existing LVLMs. In the first stage, we employ Group Relative Policy Optimization (GRPO) with template and answer-based rewards to cultivate structured, step-by-step reasoning. In the second stage, we conduct online preference optimization, where the model samples reasoning paths from GRPO-generated data, self-evaluates them, and aligns itself toward the preferred trajectories. Incorrect or suboptimal paths are concurrently stored in a Negative Replay Memory (NRM) as hard negatives, which are periodically revisited to prevent the model from repeating prior mistakes and to facilitate continual reasoning refinement. Extensive experiments show that \ours\ effectively prunes invalid reasoning paths, substantially enhances reasoning accuracy (with 7.4\% improvements on average), and yields more stable and consistent chains of thought. 
Our code will be available at \url{https://github.com/aiming-lab/PSO}.
\end{abstract}    
\section{Introduction}
\label{sec:intro}
Large Vision-Language Models (LVLMs) have demonstrated remarkable capabilities in cross-modal understanding ~\cite{liu2023visual,li2024llavaonevisioneasyvisualtask,li2022blipbootstrappinglanguageimagepretraining,hurst2024gpt,team2023gemini}. The integration of Chain-of-Thought (CoT) reasoning has further extended their applicability to complex tasks such as mathematical problem-solving ~\cite{shi2024mathllavabootstrappingmathematicalreasoning, hu2025mllmsabsorbmathreasoning, yang2024mathglmvisionsolvingmathematicalproblems, xiang2025atomthinkmultimodalslowthinking}, logical reasoning ~\cite{yan2025positionmultimodallargelanguage,huang2025visionr1incentivizingreasoningcapability,wang2025visionekiplexternalknowledgeinfusedpolicy}, and spatial reasoning ~\cite{Liu_2025,xu2025multispatialmllmmultiframespatialunderstanding,ma2025spatialllmcompound3dinformeddesign}, enabling models to decompose challenging problems into structured sequences of reasoning steps.
\begin{figure}[t!]
    \centering
    \includegraphics[width=0.45\textwidth]{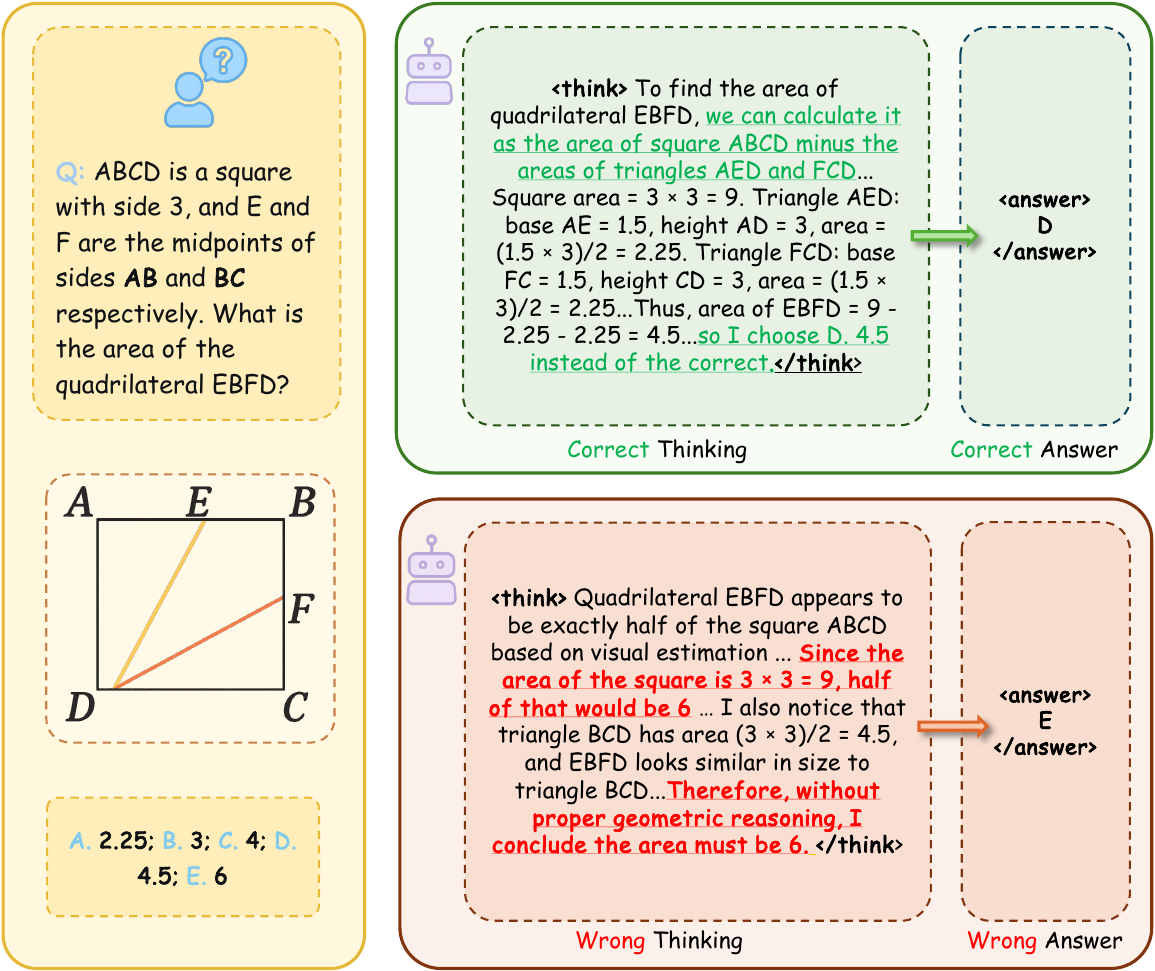} 
    \caption{LVLMs can output coherent yet flawed reasoning, but for the same question may produce correct chains, revealing that these errors arise from unstable reasoning rather than inability.}
    \label{fig:fail_case_1}
    \vspace{-4mm}
\end{figure}
Despite this progress, a critical limitation remains: when sampling reasoning paths, LVLMs often produce trajectories that appear coherent on the surface but contain latent logical errors or systematic flaws, ultimately leading to incorrect answers. Yet, for the same question, the model can occasionally generate valid reasoning paths that yield the correct solution. This observation suggests that the model often possesses the necessary knowledge, but its reasoning failures primarily stem from instability in the reasoning process (Fig.~\ref{fig:fail_case_1}).

\begin{figure*}[t!]
    \centering
    \includegraphics[width=0.90\linewidth]{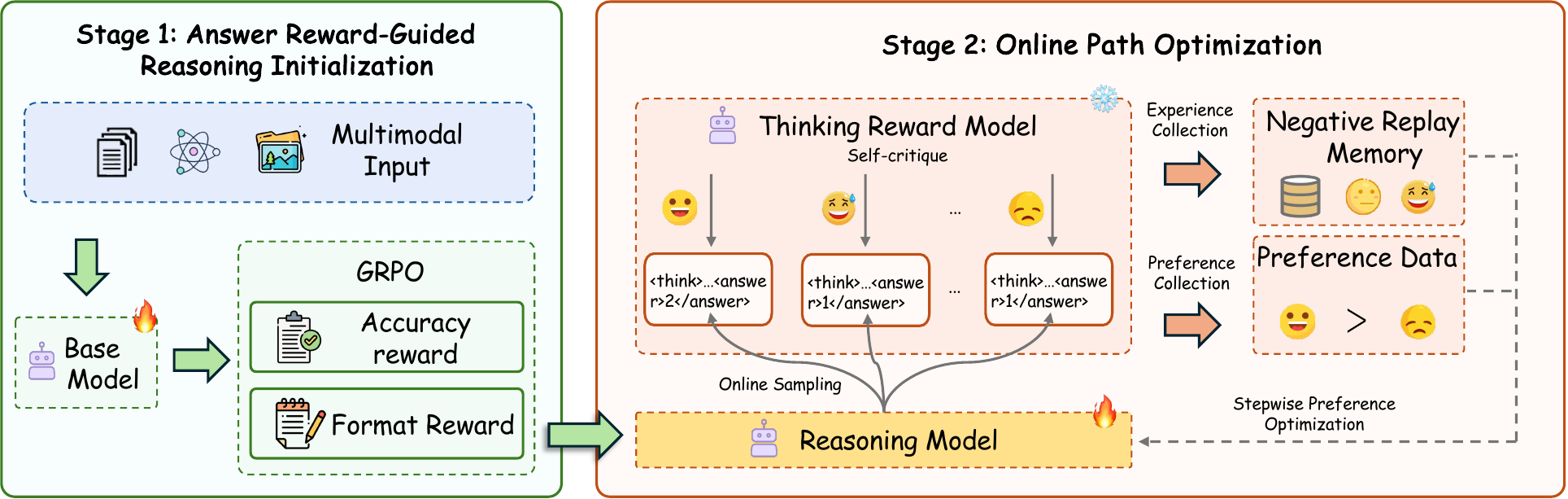}
\caption{Overview of \ours. Stage 1 (Answer Reward-Guided Reasoning Initialization): starting from a base LVLM, GRPO with accuracy and format rewards on multimodal inputs initializes structured step-by-step reasoning. Stage 2 (Online Path Optimization): for each query, the model samples multiple think–answer trajectories, a thinking reward model ranks them, and low-quality paths are stored in Negative Replay Memory as hard negatives for online, on-policy preference optimization. This closed loop prunes brittle paths and shifts probability toward stable, reliable reasoning trajectories.}
    \label{fig:algo}
    \vspace{-4mm}
\end{figure*}

This gap between potential and realized performance reveals a more fundamental issue: current LVLMs do not inherently lack reasoning capacity but may biased toward selecting unstable or flawed trajectories from the space of seemingly plausible solutions.
%HY.11.13: 这个地方最好有一张图
However, mainstream post-training paradigms still operate primarily at the answer level. Reinforcement learning methods such as Group Relative Policy Optimization (GRPO) ~\cite{guo2025deepseek} typically reward only final correctness, making it difficult to distinguish brittle reasoning chains that coincidentally yield correct answers from genuinely rigorous reasoning processes. Consequently, such methods struggle to suppress fragile reasoning behaviors. Preference optimization approaches, including Direct Preference Optimization (DPO)~\cite{rafailov2023direct}, attempt to align model outputs with human preferences, but usually rely on static preference datasets that cannot adapt to the continuously evolving model distribution during training, leading to a persistent mismatch between the model’s current reasoning patterns and its alignment signals~\cite{wang2025multimodal}.

To directly mitigate path selection bias and stabilize the reasoning process of LVLMs, we propose \ours\ (Figure~\ref{fig:algo}), a two-stage post-training framework that operates at the reasoning level rather than solely at the answer level. In the first stage, we apply GRPO with both template reward and answer reward to encourage structured, step-by-step reasoning behaviors and provide a more stable initialization for subsequent path-level optimization. In the second stage, we perform online preference optimization over reasoning paths: for each problem, we continuously sample multiple reasoning trajectories from the current policy, let the model self-assess their logical quality, and select high-quality paths as immediate positive alignment signals; meanwhile, incorrect or non-preferred trajectories are stored in a dedicated Negative Replay Memory (NRM) and repeatedly sampled as hard negatives in subsequent training iterations, explicitly penalizing previously observed flawed patterns. Unlike traditional approaches that rely on static preference datasets, this online mechanism keeps path-level supervision signals co-evolving with the model’s current reasoning distribution, effectively alleviating the mismatch between static data and a dynamically changing model. Through this design, \ours\ fully exploits the model’s diverse reasoning attempts, prunes unstable paths, and progressively concentrates probability mass on consistent and logically reliable chains of thought, thereby substantially improving both reasoning accuracy and stability.

% Building on the above observations, we characterize the root cause of unstable LVLM reasoning through the lens of path selection bias, and propose a path-level optimization solution0.

The primary contributions of this paper are threefold: (1) we show, via the substantial gap between Pass@K and Pass@1 and diverse case analyses, that the errors of existing LVLMs arise from instability in reasoning paths rather than knowledge deficiency, and formally introduce path selection bias as a key underlying problem; (2) we propose \ours, a two-stage post-training framework that combines GRPO-based initialization with online path-level preference optimization, where self-evaluation and an NRM-based negative replay mechanism continuously correct flawed reasoning patterns during training, shifting probability mass from brittle paths toward stable ones; (3) extensive experiments on diverse multimodal reasoning benchmarks demonstrate that \ours\ significantly improves answer accuracy and reasoning stability, while consistently enhancing path quality, robustness, and interpretability, providing a general and efficient paradigm for building reliable and trustworthy multimodal reasoning systems.
%第三点这里要写一点，比如提升了多少，具体数字

\section{Preliminaries}
\label{sec:preliminaries}
In this section, we review the foundations of preference-based alignment that underpin our approach. Direct Preference Optimization (DPO) ~~~\cite{rafailov2023direct} has emerged as a lightweight alternative to reinforcement learning with human feedback (RLHF) for aligning LVLMs with human preferences. Unlike RLHF, DPO directly optimizes model parameters using preference pairs without the need for an explicit reward model.
Given a prompt $x$ and two candidate responses $(y_{w}, y_{l})$ labeled as preferred and dispreferred, DPO optimizes the policy $\pi_\theta$ by encouraging higher likelihood of $y_{w}$ relative to $y_{l}$, using a logistic loss derived from the Bradley-Terry model:  
\begin{equation}
\label{equ:dpo}
\begin{split}
\mathcal{L}_{\text{DPO}}(\pi_{\theta};\pi_{\text{ref}})
= &- \mathbb{E}_{(x, y_w, y_l) \sim \mathcal{D}} \Big[
\log \sigma \Big(
\beta \log \frac{\pi_{\theta}(y_w \mid x)}{\pi_{\text{ref}}(y_w \mid x)}
\\
&\quad - \beta \log \frac{\pi_{\theta}(y_l \mid x)}{\pi_{\text{ref}}(y_l \mid x)}
\Big)
\Big],
\end{split}
\end{equation}
where $\sigma(\cdot)$ denotes the sigmoid function and $\beta$ controls the sharpness of preference distinction. 
% This formulation provides stable training and avoids high-variance policy gradient updates.

Despite its simplicity, conventional DPO is typically performed offline with static preference datasets, most of which are either manually annotated or distilled from closed-source models such as GPT or Gemini.
This introduces two limitations: (1) the feedback cannot adapt to the evolving distribution of the model during training, and (2) preference data is often generated by a different model, leading to an off-policy mismatch. 
To overcome these issues, we adopt an online, on-policy variant of DPO that updates the policy at each step using freshly sampled trajectories and self-evaluation signals: the model continuously generates candidate reasoning paths, scores them with a self-reward mechanism, and updates its preferences accordingly. 
%HY.11.13: 这里估计需要加一些self-rewarding的citation之类的，这个很多paper。

% We provide the detailed formulation of this procedure in Section~\ref{sec:method}.
\section{Path-Select Optimization}
\label{sec:method}

\subsection{Overview}
\label{sec:overview}
In this section, we present \ours, a two-stage post-training framework that directly optimizes LVLMs at the level of reasoning paths to mitigate path selection bias (Fig.~\ref{fig:algo}). Stage I, Answer Reward-Guided Reasoning Initialization, starts from a base LVLM and applies Group Relative Policy Optimization (GRPO) using accuracy and format rewards on multimodal inputs to cultivate structured, step-by-step think–answer trajectories, providing a stable initialization for subsequent path-level learning (\ref{sec:grpo}). Stage II, Online Path Optimization, repeatedly samples multiple reasoning trajectories for each query from the current policy, evaluates their logical quality with a thinking reward model to construct preference signals (\ref{sec:outcom_reward}), and updates the model via online, on-policy preference optimization (\ref{sec:online_dpo}). Low-quality trajectories are stored in a Negative Replay Memory as hard negatives (\ref{sec:memory}) and are periodically replayed to explicitly penalize brittle patterns. This closed-loop design keeps supervision synchronized with the model’s evolving behavior, prunes unstable reasoning paths, and progressively concentrates probability mass on consistent, reliable chains of thought. We detail each component in the following subsections, and summarize the overall procedure in Algorithm~\ref{algo : pso}.
\subsection{Answer Reward-Guided Reasoning Initialization}
\label{sec:grpo}
To provide a stable starting point for later path-level optimization, Stage I uses Group Relative Policy Optimization (GRPO) to guide the model toward producing clean and structured think–answer reasoning traces. During training, the model is supervised with two simple rewards: a format reward that encourages consistent separation of the reasoning process and the final answer, and an answer reward that checks whether the predicted answer matches the ground truth using rule-based evaluation for numerical and multiple-choice questions. For each input, the model samples multiple reasoning paths, and GRPO increases the likelihood of high-reward trajectories by comparing their relative quality within the group. The optimization objective is defined as follows:

\begin{align}
\mathcal{J}_{\mathrm{GRPO}}(\theta) &=
\mathbb{E}_{x \sim D,\ \{o_i\}_{i=1}^G \sim \pi_{\theta_{\text{old}}}(o|x)} \notag \\
\bigg[
\frac{1}{G} \sum_{i=1}^G &
\frac{\pi_{\theta}(o_i|x)}{\pi_{\theta_{\text{old}}}(o_i|x)} A_i
- \beta\, \mathbb{D}_{\mathrm{KL}}(\pi_{\theta} \| \pi_{\mathrm{ref}})
\bigg] 
\end{align}
where $G$ is the group size, $\hat{A}_i$ denotes the advantage estimate for the $i$-th trajectory computed from the composite reward $R_i = R_{\text{format}} + R_{\text{answer}}$, $\beta$ controls the strength of the KL regularization against the reference policy $\pi_{\text{ref}}$, and $D_{\text{KL}}(x) = \text{KL}[\pi_\theta(\cdot|x) \| \pi_{\text{ref}}(\cdot|x)]$. This stage does not aim to refine the logical structure of reasoning but instead establishes stable, organized, and easy-to-optimize trajectories, forming a strong initialization for the subsequent online path-level optimization in Stage II.  

%这个地方加一个section，让这个完整一点，可以写的比较简单，给一个grpo的公式啥的，简单描述一下。

\subsection{Online Path Optimization}
In this subsection, we describe Stage II of PSO, the online path optimization phase, outlining the reasoning-aware reward formulation, the negative replay-memory design, and the overall optimization pipeline.
\subsubsection{Reasoning-Aware Reward}
\label{sec:outcom_reward}
We first introduce the reasoning-aware reward for path-preference selection, which provides fine-grained supervision by combining rule-based outcome rewards with process-level thinking rewards. While outcome rewards ensure the correctness of final answers, thinking rewards explicitly assess the coherence and quality of intermediate reasoning steps, thereby encouraging the model to generate logically sound and interpretable reasoning traces.

\noindent \textbf{Rule-based Outcome Rewards.} Following DeepSeek-R1~\cite{guo2025deepseek}, we employ rule-based outcome reward \(R_{o} \) to generate supervision signals for each query. These functions are tailored to specific task types and evaluate model outputs by comparing them against reference answers. To ensure reliable outcome evaluation, Our training data comprise: (1) Numerical Tasks: A binary reward is assigned, with a score of 1 for exact matches between predicted and reference values, and 0 otherwise; (2) Multiple-Choice Tasks: The reward is determined by whether the predicted option corresponds to the correct choice.

\noindent \textbf{Thinking Rewards.} To enable fine-grained evaluation of reasoning quality in LVLMs, we introduce the thinking reward, a self-rewarding mechanism that leverages the base model to assign a score \(R_{t} \in [0,1]\) based solely on the quality of the intermediate reasoning process, independent of the correctness of the final answer. For a given query \(q\) and a model-generated reasoning path \(r\), the model employs a structured prompt template (see Appendix \ref{appendix:prompt_thinking}) to ensure standardized and reproducible self-assessment. 

\subsubsection{Negative Replay Memory}
\label{sec:memory}
\begin{algorithm}[t!]
\caption{Path-Select Optimization (PSO)}
\label{alg:pso}
\begin{algorithmic}[1]
\Require Policy model $\pi_{\theta}$; memory bank $\mathcal{M}$; dataset $D$; rewards $R_o$, $R_t$; parameters $\lambda$, $\tau$, $C$, $E$.
\Ensure Optimized policy $\pi_{\theta^*}$
\For{$ \text{epoch} = 1 $ to $ E $}
    \For{each $ x \in D $}
        \State Sample problem $x \sim D$ and retrieve $n$ lowest-reward traces from $\mathcal{M}(x)$ to form augmented prompt $p_{\text{aug}}$.
        \State Generate $G$ reasoning chains $\{y_i\}_{i=1}^G \sim \pi_{\theta}(\cdot|p_{\text{aug}})$.
        \State Compute composite rewards $R_i = \lambda R_t(y_i) + (1-\lambda)R_o(y_i)$.
        \State Select preferred and dispreferred responses: $y_w=\arg\max_i R_i$, $y_l=\arg\min_i R_i$.
        \If{$R_i < \tau$} 
            \State Store $(x, y_i, R_i)$ into $\mathcal{M}$ and remove oldest if $|\mathcal{M}(x)|>C$.
        \EndIf
        \State Compute DPO loss $\mathcal{L}_{\text{DPO}}$ using $(x, y_w, y_l)$ and update $\pi_{\theta}$ (Eq.~\ref{equ:dpo}).
    \EndFor
\EndFor
\end{algorithmic}
\label{algo : pso}
\end{algorithm}

While the reasoning-aware reward provides fine-grained supervision over reasoning quality, 
it cannot fundamentally prevent the model from repeating the same logical mistakes.
To explicitly address this issue, 
we introduce a Negative Replay Memory (NRM) that records and reuses low-quality reasoning samples, 
helping the model ``remember its errors'' and learn to avoid them during subsequent updates.

For each input sample $x$, 
the model generates $G$ reasoning paths $\{y_i\}_{i=1}^{G}$, 
each evaluated by a composite reward:
\begin{equation}
R_i = \lambda R_t(y_i) + (1 - \lambda) R_o(y_i),
\label{eq:composite_reward}
\end{equation}
where $R_t$ denotes the thinking reward assessing the reasoning process, 
$R_o$ denotes the outcome reward reflecting answer correctness, 
and $\lambda \in [0,1]$ balances the two signals.

We treat samples with reward $R_i < \tau$ as \textit{error cases}.
For each sample $x$, 
the lowest-reward trajectory is stored in the memory bank $\mathcal{M}$:
\begin{equation}
\mathcal{M}(x) \leftarrow \text{append}\big(y_i, R_i\big), \quad \text{if } R_i < \tau.
\label{eq:memory_append}
\end{equation}
When the number of stored entries for $x$ exceeds the capacity $C$, the oldest item is discarded following a FIFO policy.
In this way, $\mathcal{M}$ continuously maintains a compact, up-to-date record of each sample’s most recent reasoning failures, ensuring that the memory remains aligned with the model’s evolving capabilities during training.

Building on this memory mechanism, the model then leverages these historical mistakes during subsequent epochs. When the same sample $x$ reappears, 
the model retrieves $n$ lowest-reward reasoning paths $\{y^{-}_j\}_{j=1}^{n}$ from $\mathcal{M}(x)$.
These historical negatives are embedded into the input prompt following the template described in Appendix \ref{memory_prompt}.
By explicitly incorporating its past failures into the current reasoning context, the model performs contrastive self-reflection, comparing new reasoning paths against previously incorrect ones. Through repeated exposure to these structured contrasts, the model gradually develops more stable, consistent, and logically coherent reasoning behaviors while preserving sample-level continuity across training.
\begin{figure*}[t!]
    \centering
    \begin{subfigure}{0.33\textwidth}
        \centering
        \includegraphics[width=\linewidth]{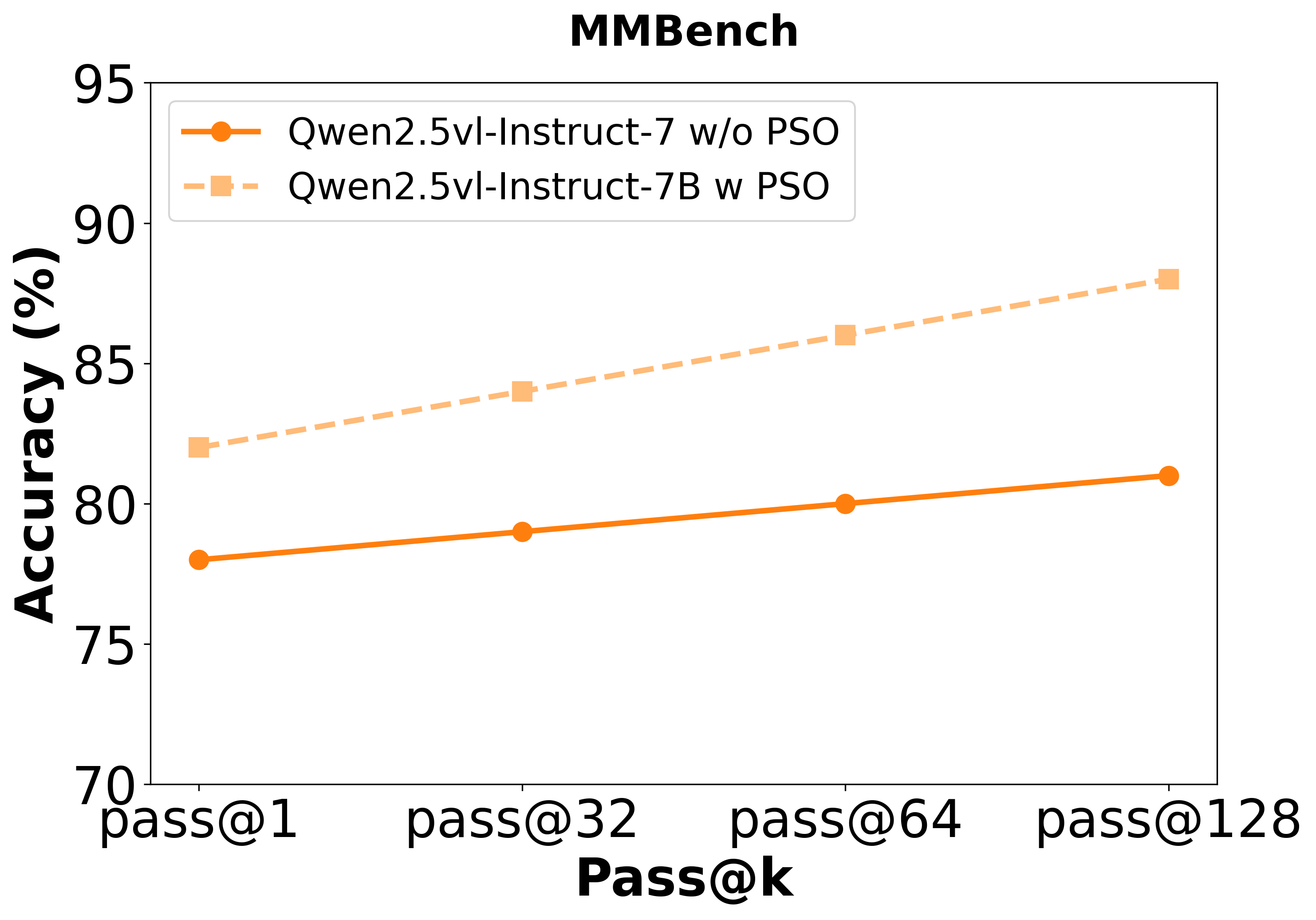}
    \end{subfigure}
    \hfill
    \begin{subfigure}{0.33\textwidth}
        \centering
        \includegraphics[width=\linewidth]{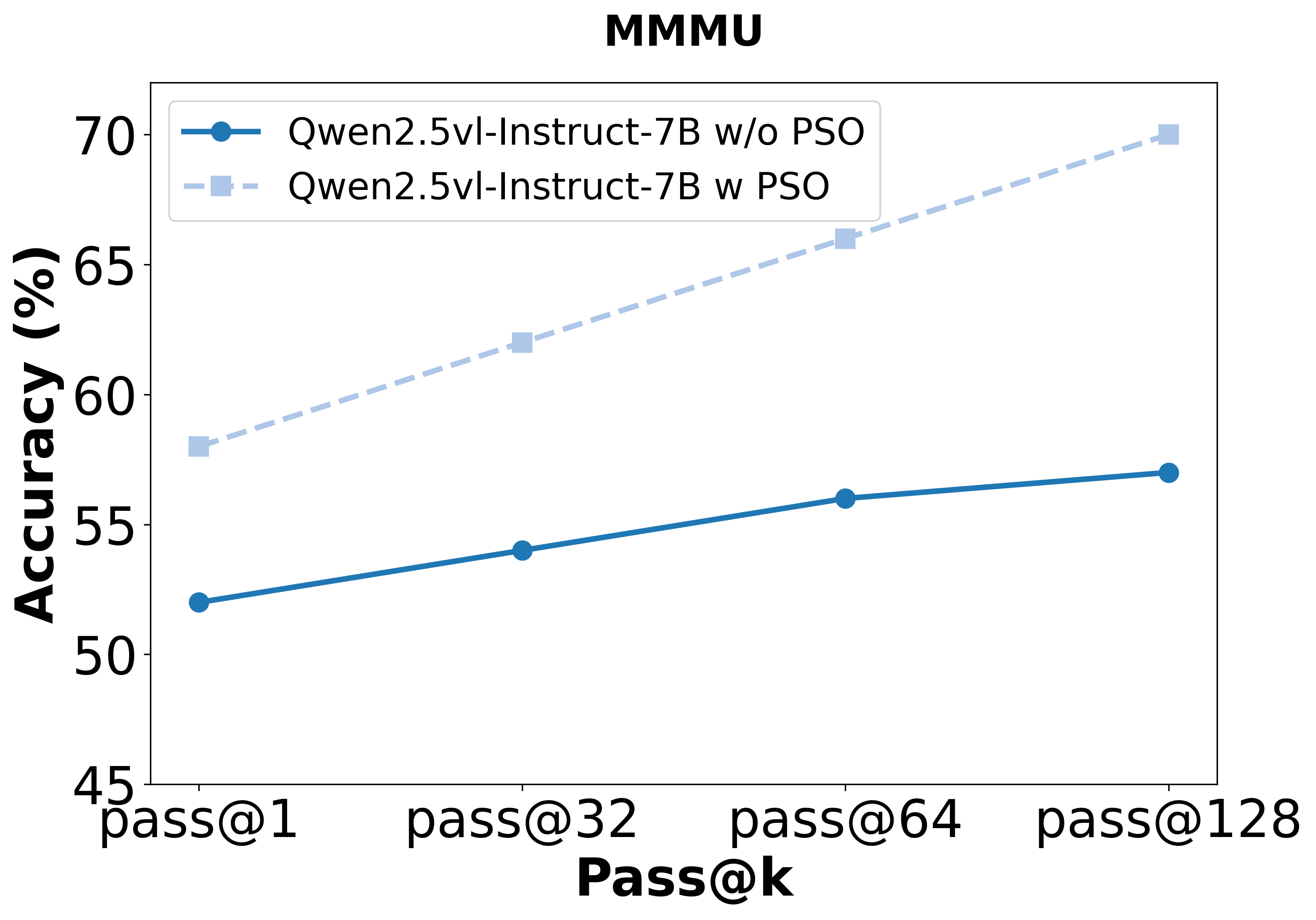}
    \end{subfigure}
    \hfill
    \begin{subfigure}{0.33\textwidth}
        \centering
        \includegraphics[width=\linewidth]{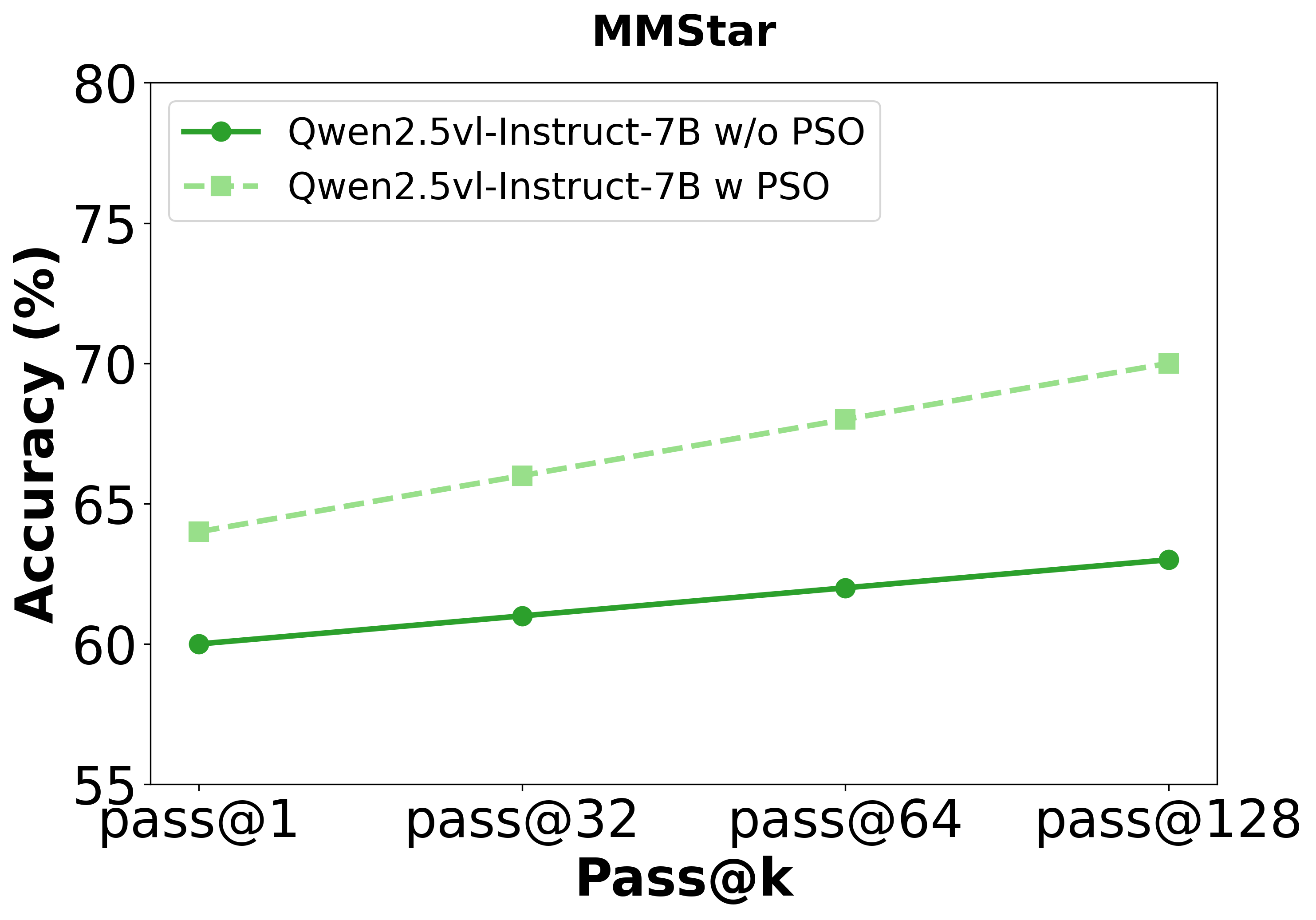}
    \end{subfigure}
    \caption{Comparison of Pass@k performance on MMBench and MMBench-Star for Qwen2.5VL-7B-Instruct before and after PSO.}
    \label{fig:three_benchmarks}
\end{figure*}
\subsubsection{Online Preference Optimization with Memory Retrieval}
%HY.11.13: 这个标题不是很好，可以改一下
\label{sec:online_dpo}
Building upon NRM, 
we further integrate it into an online preference optimization framework, 
enabling real-time adaptation and error-driven refinement. Specifically, for a given input $x$, 
the model retrieves $n$ lowest-reward responses $\{y^{-}_i\}_{i=1}^{n}$ from $\mathcal{M}(x)$ 
and incorporates them into an augmented prompt $p_{\text{aug}}$. 
Using this prompt, 
the current policy $\pi_{\theta}$ samples $G$ new reasoning paths $\{y_i\}_{i=1}^{G}$, 
each scored using Eq.~\ref{eq:composite_reward}. 
We then select the highest-reward path $y_{w}$ and lowest-reward path $y_{l}$ 
to form a preference pair $(y_{w}, y_{l})$ and update the policy via Eq.~\ref{equ:dpo}.
\begin{table*}[t!]
\centering
\small
\caption{Comparison of models on MathVista and MathVerse. 
The best is \textbf{bold}, and the runner-up is \underline{underline}.
\textsuperscript{1}Scientific Reasoning (SCI), 
\textsuperscript{2}Textbook Question Answering (TQA), 
\textsuperscript{3}Arithmetic Reasoning (ARI), 
\textsuperscript{4}Math Word Problem (MWP), 
\textsuperscript{5}Logical Reasoning (LOG), 
\textsuperscript{6}Vision Intensive (VI), 
\textsuperscript{7}Vision Only (VO), 
\textsuperscript{8}Vision Dominant (VD), 
\textsuperscript{9}Text Dominant (TD), 
\textsuperscript{10}Text Lite (TL).}
\resizebox{\linewidth}{!}{%
\renewcommand{\arraystretch}{1.05}
\begin{tabular}{l|*{6}{c}|*{6}{c}}
\toprule
\multirow{2}{*}{\textbf{Model}} & \multicolumn{6}{c|}{\textbf{MathVista}} & \multicolumn{6}{c}{\textbf{MathVerse}} \\
\cmidrule{2-13}
 & \textit{\textbf{AVG}} & SCI\textsuperscript{1} & TQA\textsuperscript{2} & ARI\textsuperscript{3} & MWP\textsuperscript{4} & LOG\textsuperscript{5} & \textit{\textbf{AVG}} & VI\textsuperscript{6} & VO\textsuperscript{7} & VD\textsuperscript{8} & TD\textsuperscript{9} & TL\textsuperscript{10} \\
\midrule
\multicolumn{13}{l}{\textit{Open-Source General MLLMs}} \\
LLaVA-OneVision-7B~\cite{li2024llava} & 63.2 & 65.6 & 60.8 & 57.8 & 69.4 & 21.6 & 26.2 & - & - & - & - & - \\
LLaVA-OneVision-72B~\cite{li2024llava} & 68.4 & 63.1 & 65.8 & 60.1 & 73.7 & \underline{27.1} & 27.2 & - & - & - & - & - \\
Cambrian-1-34B~\cite{tong2024cambrian} & 50.9 & 53.3 & 55.1 & 45.6 & 51.6 & 16.2 & - & - & - & - & - & - \\
GPT-4V & 51.8 & 63.1 & 65.8 & 51.8 & 57.5 & 21.6 & 32.8 & - & - & - & - & - \\
\midrule
\multicolumn{13}{l}{\textit{Open-Source Math MLLMs}} \\
Math-LLaVA-13B~\cite{shi2024math} & 46.6 & 49.2 & 51.3 & 40.2 & 56.5 & 16.2 & 22.9 & 24.5 & 16.1 & 21.7 & 27.3 & 24.9 \\
Math-PUMA-Qwen2vl-7B~\cite{zhuang2025math} & 47.9 & 42.6 & 46.2 & 46.2 & 68.3 & 21.6 & 33.6 & 33.4 & 26.0 & 31.6 & 42.1 & 35.0 \\
Multimath-7B~\cite{peng2024multimath} & 50.0 & - & 50.0 & - & 61.8 & - & 26.9 & 28.1 & 15.0 & 25.9 & 34.8 & 30.8 \\
URSA-8B~\cite{luo2025ursa} & 59.8 & 58.2 & 63.9 & 53.5 & \underline{75.3} & 21.6 & 45.7 & \textbf{46.4} & 34.6 & \underline{43.9} & 55.3 & \underline{48.3} \\
\midrule
\multicolumn{7}{l}{\textit{Open-Source Reasoning MLLMs}} \\
Curr-ReFT-7B~\cite{deng2025boosting} & 64.5 & - & - & - & - & - & - & - & - & - & - & - \\ 
R1-OneVision-7B~\cite{yang2025r1} & 64.1 & 61.5 & 62.0 & 56.1 & 64.5 & 16.2 & \underline{46.4} & - & 40.0 & - & - & - \\
InternVL2.5-8B-VisualPRM~\cite{wang2025visualprm} & 68.5 & 61.5 & 53.9 & 45.9 & 66.8 & 21.2 & 30.7 & 28.9 & 35.8 & 27.3 & 31.7 & 29.7 \\ 
\midrule
Qwen2.5vl-Instruct-7B~\cite{bai2025qwen2} & 67.5 & 65.6 & 67.7 & 57.5 & 69.4 & 27.0 & 44.0 & 41.1 & 41.0 & 38.7 & 55.2 & 44.0 \\
\quad +SFT+GRPO & \underline{69.5} & \underline{69.4} & \underline{72.5} & \textbf{60.8} & 70.2 & 23.6 & 45.8 & 39.4 & \underline{41.2} & 41.4 & \underline{55.5} & 45.3 \\
\rowcolor{gray!25} \quad \textbf{+\ours\ (Ours)}\ & \textbf{70.8} & \textbf{70.3} & \textbf{72.9} & \underline{59.4} & \textbf{76.5} & \textbf{35.7} & \textbf{47.6} & \underline{45.7} & \textbf{43.9} & \textbf{44.8} & \textbf{58.6} & \textbf{51.0} \\
\bottomrule
\end{tabular}}
\label{tab:math_results}
\end{table*}
This online preference optimization procedure, augmented with memory retrieval, provides two primary benefits. (1) On-policy adaptivity. All reasoning samples are generated from the current policy, keeping preference signals aligned with the model’s evolving output distribution and avoiding the distribution mismatch common in offline preference datasets.
(2) Error-aware refinement. Incorporating hard negatives from $\mathcal{M}$ enables explicit comparison between successful and failed reasoning paths, strengthening logical consistency and discouraging repeated error modes. Taken together, NRM and online DPO constitute a closed-loop training process: the model iteratively generates trajectories, evaluates them, retrieves prior failures, and updates its policy based on contrastive preferences. This cycle encourages the policy to allocate more probability mass to coherent and stable reasoning paths while progressively suppressing unreliable ones.
% Together, the NRM and online DPO form a closed-loop optimization cycle: 
% the model continually samples, evaluates, replays, and refines its reasoning trajectories. 
% This self-corrective learning paradigm progressively shifts probability mass 
% toward stable and reliable chains of thought, 
% realizing a form of autonomous reasoning refinement.

\section{Experiment} \label{sec:experiment}
In this section, we demonstrate the effectiveness of \ours\ by addressing three key questions: (1) Does it improve performance across both mathematical and general multimodal benchmarks? (2) Beyond GRPO's fixed search space, can online preference optimization post-training the model toward producing better reasoning paths? (3) Does the integration of reasoning-aware reward signals genuinely enhance both process-level (thinking) and answer-level rewards? (4) Can the Negative Replay Memory (NRM) prevent repeated failure modes by reusing incorrect or low-reward reasoning paths as hard negatives?
\begin{figure}[t!]
  \centering
  \includegraphics[width=0.9\linewidth]{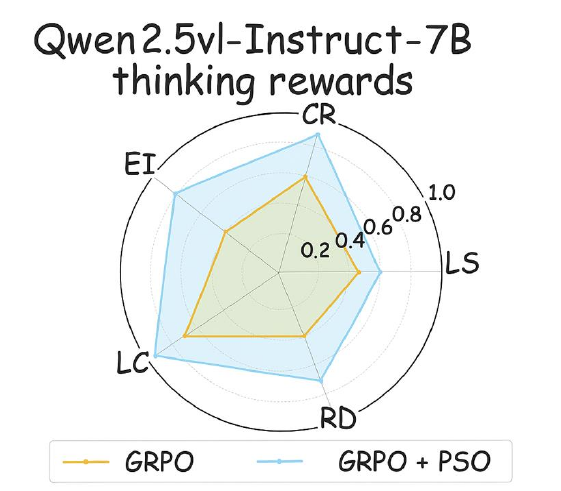}
  \caption{Qwen2.5vl-Instruct-7B thinking rewards on MMMU dataset. LS=Logical Soundness, EI = Error Identification, CR = Correct Reasoning, LC = Language Consistency, RD = Redundancy.}
  \label{fig:radar_reward}
\vspace{-6mm}
\end{figure}
\subsection{Experimental Settings} 
\label{sec:setting}
\textbf{Benchmarks.} We evaluate our model on both multimodal mathematical reasoning and general multimodal reasoning benchmarks. For mathematical reasoning, we report detailed results on MathVista~\cite{lu2023mathvista} and MathVerse~\cite{zhang2024mathverse}. For general multimodal capabilities, we conduct evaluations on MMMU~\cite{yue2024mmmu}, MME~\cite{liang2024survey}, MMStar~\cite{chen2024we}, ChartQA~\cite{masry2022chartqa}, RealWorldQA~\cite{ai2024grok} and MMBench~\cite{xu2023mmbench}.

\noindent \textbf{Implementation Details.}
We first pre-train Qwen2.5vl-Instruct-7B on the SophiaVL-R1-130k dataset~\cite{fan2025sophiavl} using SFT (cold start) + GRPO, obtaining a base model with preliminary reasoning capabilities. Building upon this initialization, we further apply the proposed online path optimization algorithm and continue training on the SophiaVL-R1-130k dataset. The training phase is conducted on 8 NVIDIA H100 80GB GPUs for a total of 550 steps, implemented within the TRL~\cite{vonwerra2022trl} framework. The training configuration includes a group size of 8, a learning rate of $5\times10^{-7}$, and 3 epochs. The sample size $K$ of online DPO is set to 8. During evaluation, we adopt the default prompts and generate responses using greedy decoding. Additional evaluation details refer to Appendix A.
\begin{table*}[t!]
\centering
\small
\caption{Comparison of models on general ability benchmarks. Best results are in \textbf{bold}, runner-ups are \underline{underlined}.}
\label{tab:general_results}
\setlength{\tabcolsep}{4pt}
\begin{tabular}{@{}lcccccc@{}}
\toprule
\textbf{Model} & \textbf{MMMU} & \textbf{MME} & \textbf{ChartQA} & \textbf{MMBench} & \textbf{MMStar} & \textbf{RealwordQA} \\
\midrule
\multicolumn{7}{l}{\textit{Open-Source General MLLMs}}
\\
LLaVA-OneVision-7B~\cite{li2024llava} & 48.8 & 1998.0 & 80.0 & -- & 61.7 & 66.3\\
LLaVA-OneVision-72B~\cite{li2024llava} & 56.8 & 2261.0 & 83.7 & -- & \underline{66.1}  & 71.9 \\
Cambrian-1-34B~\cite{tong2024cambrian} & 49.7 & 1689.3 & 75.6 & 81.4 & 54.2 & --\\
GPT-4V & 56.8 & 1926.0 & 78.5 & 75.0 & 57.1 & 61.4 \\
\midrule
\multicolumn{7}{l}{\textit{Open-Source Math MLLMs}} \\
URSA-8B~\cite{luo2025ursa} & 43.1 & 1605.7 & 44.4 & 55.5 & 42.3 & -- \\
\midrule
\multicolumn{7}{l}{\textit{Open-Source Reasoning MLLMs}} \\
Curr-ReFT-7B~\cite{deng2025boosting} & -- & -- & -- & 79.0 & -- & -- \\
R1-Onevision-7B~\cite{yang2025r1} & 51.6 & 2223.3 & -- & 75.6 & 59.1 & -- \\
InternVL2.5-8B-VisualPRM~\cite{wang2025visualprm} & 56.2 & -- & 60.8 & 83.5 & 63.4 & -- \\
\midrule
Qwen2.5vl-Instruct-7B~\cite{bai2025qwen2} & 57.4 & 2306.0 & 86.3 & 83.3 & 64.3 & 68.5 \\
\quad + SFT + GRPO & \underline{58.7} & \underline{2343.0} & \underline{89.1} & \underline{85.1} & 64.8 & \underline{70.2} \\
\rowcolor{gray!25}
\quad \textbf{+ \ours\ (Ours)} & \textbf{60.9} & \textbf{2376.7} & \textbf{91.2} & \textbf{86.4} & \textbf{66.5} & \textbf{72.3} \\
\bottomrule
\end{tabular}
\end{table*}
\begin{table*}[t!]
\centering
\caption{Ablation study of \ours\ across benchmarks, showing performance drops when removing thinking reward, online DPO, or memory retrieval.}
\label{tab:ablation}
\small
\setlength{\tabcolsep}{2.3pt}
\begin{tabular}{lcccccccc}
\toprule
\textbf{Model} & \textbf{MathVista} & \textbf{MathVerse} & \textbf{MMMU} & \textbf{MME} & \textbf{ChartQA} & \textbf{MMBench} & \textbf{MMStar} & \textbf{RealwordQA} \\
\midrule
Qwen2.5vl-Instruct-7B + SFT + GRPO & 69.5 & 45.8 & 58.7 & 2343.0 & 89.1 & 85.1 & 64.8 & 70.2\\
\ours\ w/o thinking reward & 69.8 & 45.3 & 59.2 & 2359.0 & 89.5 & 85.9 & 65.1 & 71.5 \\
\ours\ w/o online DPO & 68.9 & 46.1 & 58.9 & 2350.2 & 89.8 & 85.4 & 65.0 & 70.9\\
\ours\ w/o memory retrieval & 70.0 & 46.3 & 59.0 & 2355.4 & 90.0 & 85.6 & 65.3 & 71.7\\
\rowcolor{gray!25}
Qwen2.5vl-Instruct-7B + \ours\ & \textbf{70.8} & \textbf{47.6} & \textbf{60.9} & \textbf{2376.7} & \textbf{91.2} & \textbf{86.4} & \textbf{66.5} & \textbf{72.3}\\
\bottomrule
\end{tabular}
% \vspace{-0.5cm}
\end{table*}
\subsection{Main Results}
\begin{figure}[t!]
  \centering
  \includegraphics[width=0.9\linewidth]{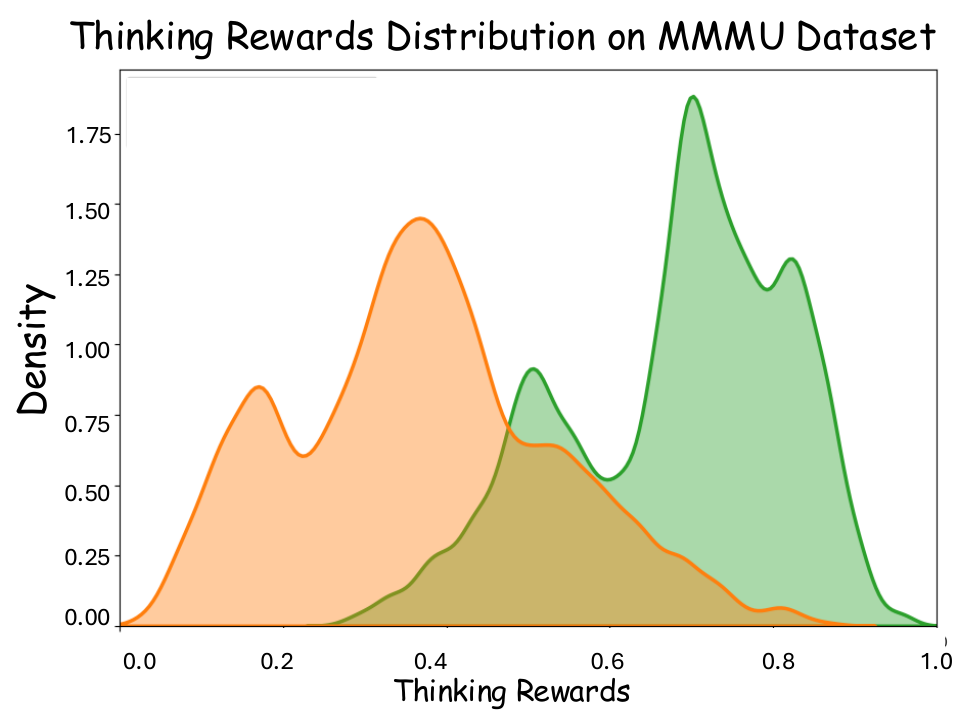}
  \caption{Reasoning reward distribution on the MMMU dataset with different methods. \textcolor{orange}{Orange} denotes the rewards of reasoning paths sampled from Qwen2.5vl-Instruct-7B, while \textcolor{green}{green} denotes those from Qwen2.5vl-Instruct-7B + \ours.}
  \label{fig:reasoning_reward}
  \vspace{-6mm}
\end{figure}

We present the results of \ours\ across mathematical reasoning benchmarks (Table~\ref{tab:math_results}), general multimodal benchmarks (Table~\ref{tab:general_results}), and post-training evaluations (Fig.~\ref{fig:reasoning_reward} and Fig.~\ref{fig:radar_reward}), showing state-of-the-art or competitive performance, strong generalization, and notable improvements in reasoning rewards. A detailed analysis is provided below.

% \begin{figure*}[t!]
%   \centering
%   \includegraphics[width=\linewidth]{author-kit-CVPR2026-v1-latex-/fig/pass@k.png}
%   \caption{Reasoning reward distribution on the MMMU dataset with Qwen2.5vl-Instruct-7B.}
%   \label{fig:reasoning_reward_xx}
% \end{figure*}

\noindent \textbf{Performance on Math Reasoning Benchmarks.}
As shown in Table~\ref{tab:math_results}, \ours\ delivers strong competitive performance on mathematical reasoning benchmarks. On MathVista, it achieves 70.8\% accuracy, notably surpassing numerous open-source reasoning models and even outperforming LLaVA-OneVision-72B with only one-tenth of its parameters. Compared with Qwen2.5vl-Instruct-7B (SFT+GRPO), \ours\ shows substantial gains up to 1.8\% on MathVerse and consistently outperforms the baseline across all sub-tasks. These results demonstrate that \ours\ can make base model leverages more effective reward signals, substantially increasing the likelihood of sampling high-quality reasoning trajectories and thereby enabling superior reasoning strategies with stronger generalization in complex tasks.

\noindent \textbf{Performance on General Benchmarks.}
Many task-specific reasoning models, particularly those optimized for mathematical problem-solving or other specialized domains, achieve strong in-domain performance but often exhibit limited generalization when evaluated on comprehensive multimodal benchmarks such as URSA-8B. In contrast, \ours\ demonstrates consistently competitive results across widely recognized general-purpose benchmarks, underscoring its superior generalization ability (Table~\ref{tab:general_results}). For instance, on the widely adopted MMMU benchmark for multi-disciplinary reasoning, \ours\ surpasses LLaVA-OneVision-72B by 4.1\%.

\noindent \textbf{Quality of Reasoning Paths After Post-training.} To assess whether \ours\ can improve the quality of the model’s reasoning paths, we take the MMMU benchmark as an example. As shown in Fig.~\ref{fig:reasoning_reward}, the post-training stage of \ours\ shifts the reward distribution of randomly sampled reasoning paths toward higher values, indicating that the model produces more coherent and consistent reasoning trajectories after \ours, which in turn directly improves answer accuracy. For a more fine-grained analysis, we further decompose the reward into sub-scores following Appendix B, with the results shown in Fig.~\ref{fig:radar_reward}. We observe that, compared with GRPO, \ours\ yields consistent improvements across multiple dimensions of reasoning quality, including logical soundness, error identification, language consistency, and correctness of reasoning. This demonstrates that post-training not only boosts outcome-level accuracy, but also enhances the robustness and interpretability of the reasoning process.

\begin{figure}[t!]
  \centering
  \includegraphics[width=0.9\linewidth]{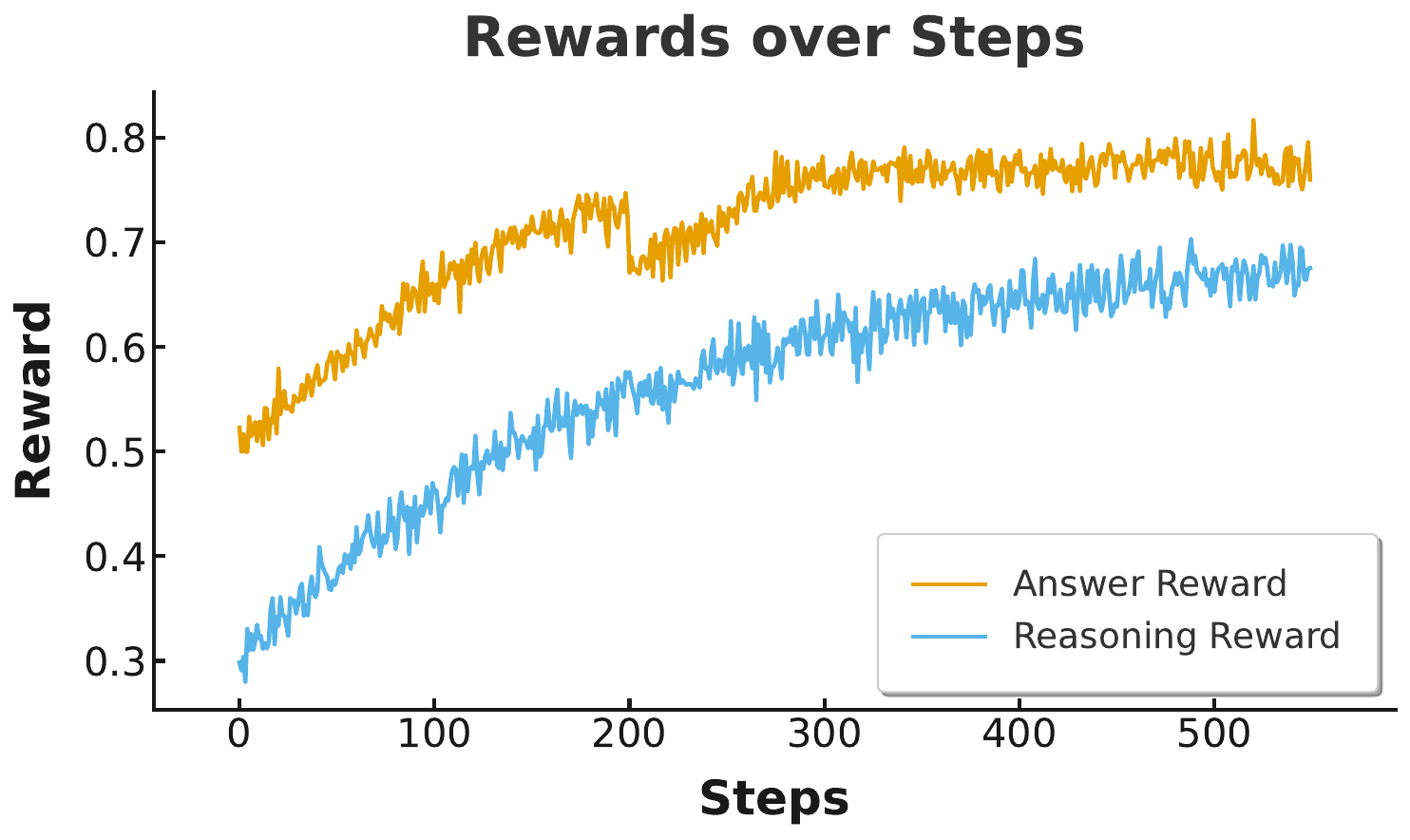}
  \caption{Answer and Reasoning Rewards as a function of training steps for Qwen2.5-VL-7B-Instruct on the MMMU dataset.}
  \label{fig:reasoning_reward_line}
  \vspace{-6mm}
\end{figure}

\noindent \textbf{Evolution of Reasoning and Answer Rewards.} As shown in Fig.~\ref{fig:reasoning_reward_line}, answer rewards rise quickly in the early stages, reflecting rapid adaptation to producing correct outputs, while reasoning rewards steadily increase throughout training, indicating progressively more coherent and logically consistent traces. This joint evolution demonstrates \ours's self-rewarding framework effectively aligns outcome accuracy with reasoning quality, ensuring post-training gains stem from deeper improvements in reasoning rather than superficial answer matching.

\subsection{Ablation Study}
In this section, we conduct ablation studies to systematically evaluate the contribution of each core component in \ours. Specifically, we examine three ablated variants: (1) \ours\ w/o thinking reward, which removes the thinking reward module while retaining the answer reward; (2) \ours\ w/o online DPO: we first use the base model to sample multiple candidates for all training examples and rank them using the reward defined in Section~\ref{sec:outcom_reward}, thereby constructing a static preference dataset. We then train the base model with DPO for 3 epochs on this fixed preference dataset, replacing the original online DPO stage; and (3) \ours\ w/o memory retrieval, which disables the memory retrieval mechanism during training. The corresponding results are reported in Table~\ref{tab:ablation}.

\noindent \textbf{Ablation of Memory Retrieval.} Disabling memory retrieval (\ours\ w/o memory retrieval) consistently reduces performance, with MathVerse accuracy dropping from 47.6 to 46.3 and MME score decreasing from 2376.7 to 2355.4. The results indicate the importance of the memory module, which replays past errors as hard negatives, prevents systematic mistakes from recurring, and enhances generalization to structurally similar problems.

% \begin{figure}[t!]
% \centering
% \includegraphics[width=\linewidth]{author-kit-CVPR2026-v1-latex-/fig/pass@k_2.png}
% \caption{Comparison of performance improvements for MMMU, MMBench, and MMStar across different training passes. Each group shows baseline (“Before”), post-training (“After”), and incremental gain ($\Delta$) scores, illustrating consistent enhancement in accuracy (\%) with increasing pass numbers.} 
% \label{fig:passk_results}
% \end{figure}
\noindent \textbf{Ablation of Online and Offline DPO.} Next, we compare online preference optimization with its offline variant. Replacing online DPO with offline DPO (\ours\ w/o online DPO) degrades MathVista accuracy from 70.8 to 68.9 and lowers general-domain results, for example reducing MMMU performance from 60.9 to 58.9. This highlights the value of online, on-policy preference sampling, which adapts to the model’s evolving reasoning patterns and mitigates the distribution shift inherent in static datasets. By contrast, offline DPO relies on outdated feedback, limiting its ability to guide reasoning effectively.

\noindent \textbf{Ablation of Thinking Reward.} We further investigate the importance of the thinking reward module. Removing it (\ours\ w/o memory retrieval) yields clear declines, with MathVerse dropping from 47.6 to 45.3 and MMStar decreasing from 66.5 to 65.1. This shows that outcome-based supervision alone is insufficient: the thinking reward provides fine-grained process-level signals, steering the model away from logically flawed but answer-correct reasoning chains, and thereby improving reasoning robustness.

\noindent \textbf{Ablation of Pass@K.} Additionally, as shown in Figure~\ref{fig:three_benchmarks}, the performance gap between PSO and the baseline widens with increasing sampling budget, evidenced by superior Pass@128 scores on both MMBench and MMStar. This trend signifies a fundamental shift in the model's reasoning strategy: PSO not only elevates the top-ranked reasoning path but systematically restructures the underlying trajectory distribution. By actively pruning flawed reasoning patterns through negative replay and reinforcing logical coherence via thinking rewards, our method increases the density of valid reasoning paths. This reduces the model's dependency on superficially plausible but ultimately brittle chains, thereby directly tackling the core issue of reasoning instability. The consistent gains across both mathematical and general-purpose benchmarks confirm that this path-level optimization effectively addresses a universal weakness in LVLMs, rather than overfitting to a specific task type.

\section{Related Work}
\label{sec:relatedwork}
\noindent \textbf{Reasoning and Alignment in LVLMs.} Recent advancements in Multimodal Large Language Models (MLLMs), such as LLaVA~\cite{liu2023visual}, Qwen-VL~\cite{bai2025qwen2}, and GPT-4V~\cite{achiam2023gpt}, have demonstrated strong performance across vision-language tasks by incorporating chain-of-thought (CoT)~\cite{wei2022chain} reasoning to decompose complex problems~\cite{wang2024stop,ge2023mllm}. However, studies consistently show that these models often produce superficially plausible yet logically flawed reasoning traces~\cite{liu2025robustness}. To address this, research has shifted from outcome-based to process-based supervision, aiming to improve both answer correctness and reasoning validity. Preference optimization methods - including RLHF~\cite{lee2023rlaif} and DPO~\cite{rafailov2023direct} have become standard for alignment, yet offline approaches suffer from distribution shift and inability to adapt to the model’s evolving state. Online DPO variants~\cite{gupta2025robust,qi2024online} have emerged to mitigate these issues through continuous sampling and evaluation. Our work extends this direction by introducing an online DPO framework~\cite{qi2024online} tailored for multimodal reasoning, integrating process-level rewards and dynamic memory to enhance robustness and sample efficiency.

\noindent \textbf{Self-Training with Reward and Critique Mechanisms.} Self-training methods leverage model-generated outputs to enable iterative improvement, employing techniques such as self-play, self-critique, and self-rewarding~\cite{zhou2025reagent, xia2024cares, xu2025towards} to reduce reliance on external supervision. Recent approaches like DeepSeek-R1~\cite{guo2025deepseek} and SophiaVL~\cite{fan2025sophiavl} utilize reinforcement learning and process rewards to guide reasoning, often depending on outcome-based rewards or heuristic rules. Our method advances this line by introducing a composite reward function that jointly evaluates reasoning quality and final answers, alongside a memory module that facilitates structured learning from past errors. This allows the model to not only improve answer accuracy but also refine its reasoning strategies over time, supporting sustained self-improvement~\cite{shi2025look}.

\noindent \textbf{Memory-Augmented Learning and Experience Replay.} The integration of memory mechanisms, such as experience replay and episodic memory, has long been used in machine learning to improve sample efficiency and combat catastrophic forgetting~\cite{huet2025episodic, zhao2024efficient, wang2025erci}. In language and reasoning models, memory-augmented transformers and retrieval-augmented generation have been employed to maintain context and support factual consistency~\cite{brown2025systematic, lampinen2025latent, morais2025general}. Our memory module draws inspiration from these architectures but is specifically designed to store and retrieve negative examples (e.g., flawed reasoning traces), thereby curbing error recurrence and supporting continuous self-refinement in multimodal reasoning tasks. This approach aligns with broader efforts to build more adaptive and resilient reasoning systems through structured memory utilization.

\section{Conclusion}
\label{sec:conclusion}
In this work, we introduce PSO, a self-rewarding direct preference optimization framework that integrates online DPO with a memory-augmented mechanism. Unlike conventional answer-supervised training, our method jointly evaluates outcome correctness and reasoning quality for fine-grained process-level alignment. The memory retrieval enhances robustness by replaying flawed reasoning traces, preventing systematic errors. Extensive experiments on mathematical and multimodal reasoning benchmarks show that PSO significantly outperforms strong baselines in both accuracy and logical consistency. Ablation studies confirm the essential roles of the thinking reward, online optimization, and memory retrieval. Overall, PSO advances multimodal LLM alignment from outcome-level supervision to reasoning-level optimization, opening promising avenues for building more interpretable and trustworthy reasoning systems.

{
    \small
    \bibliographystyle{ieeenat_fullname}
    \bibliography{main}
}

\clearpage
% \setcounter{page}{1}
% \maketitlesupplementary
\appendix

\section{Evaluation Details}
\label{appendix:eval}
Our experimental evaluations are primarily conducted using VLMEvalKit, adhering to the recommended Python package versions to ensure consistency and reproducibility. For baseline comparisons, performance metrics are sourced directly from the OpenVLM leaderboard. All evaluated models are tested with their default prompt configurations, while the answer extraction functions are adapted to align with each model’s output format. For example, in the case of R1-style models, we extract responses enclosed between the \texttt{<answer>} and \texttt{</answer>} tags.
For MathVista, evaluations are performed on the \texttt{testmini} split. For MathVerse, we report the average results across five subsets: vision-only, vision-dominant, vision-intensive, text-dominant, and text-lite. For MMMU, evaluations are carried out on the \texttt{mmmu\_dev\_val} set, while for ChartQA, assessments are based on the test set. For MMBench, we follow the standard evaluation protocol on the \texttt{MMBench\_Dev\_EN} set.

\section{Prompt Template Used for Evaluating Thinking Process Quality}
\label{appendix:prompt_thinking}
\begin{figure*}[t]
    \centering
    \includegraphics[width=\linewidth]{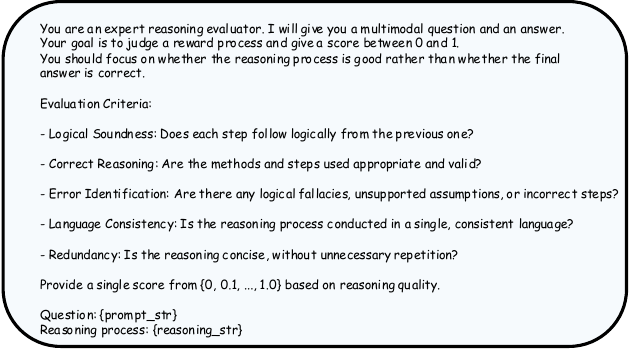}
    \caption{Prompt for evaluating thinking process quality. The evaluation criteria are derived from a systematic analysis of common error patterns in reasoning traces and are consolidated into five core dimensions: Logical Soundness, Correct Reasoning, Error Identification, Language Consistency, and Redundancy.}
    \label{fig:prompt_2}
\end{figure*}

Fig. \ref{fig:prompt_2} illustrates the prompt template for thinking process evaluation.
% \begin{tcolorbox}[colback=gray!5!white,colframe=gray!75!black,
%                    fonttitle=\bfseries,
%                   breakable]
% prompt = '''You are an expert reasoning evaluator. I will give you a multimodal question and an answer. 
% Your goal is to judge a reward process and give a score between 0 and 1. 
% You should focus on whether the reasoning process is good rather than whether the final answer is correct.

% Evaluation Criteria:

% - Logical Soundness: Does each step follow logically from the previous one?

% - Correct Reasoning: Are the methods and steps used appropriate and valid? 

% - Error Identification: Are there any logical fallacies, unsupported assumptions, or incorrect steps?

% - Language Consistency: Is the reasoning process conducted in a single, consistent language?

% - Redundancy: Is the reasoning concise, without unnecessary repetition?

% Provide a single score from \{0, 0.1, ..., 1.0\} based on reasoning quality.

% Question: \{prompt\_str\} \\
% Reasoning process: \{reasoning\_str\}

% '''
% \end{tcolorbox}

\section{Memory-Integrated Sampling Prompt Template}
\label{memory_prompt}
\begin{figure*}[t]
    \centering
    \includegraphics[width=\linewidth]{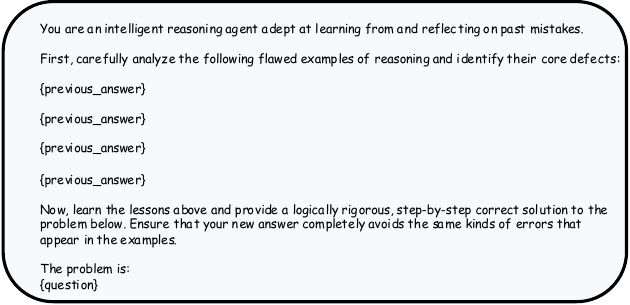}
    \caption{Memory-Integrated sampling prompt template. This prompt template integrates memory of past flawed reasoning into the query, encouraging reflection and error-avoidance so the model can generate more rigorous and robust solutions.}
    \label{fig:prompt_1}
\end{figure*}

Fig. \ref{fig:prompt_1} illustrates the memory-Integrated sampling prompt template.
% This prompt integrates memory of past flawed reasoning into the query, encouraging reflection and error-avoidance so the model can generate more rigorous and robust solutions.
% \begin{tcolorbox}[colback=gray!5!white,colframe=gray!75!black,
%                    fonttitle=\bfseries,
%                   breakable]
% prompt = '''You are an intelligent reasoning agent adept at learning from and reflecting on past mistakes.

% First, carefully analyze the following flawed examples of reasoning and identify their core defects:

% \{previous\_answer\}

% \{previous\_answer\}

% \{previous\_answer\}

% \{previous\_answer\}

% Now, learn the lessons above and provide a logically rigorous, step-by-step correct solution to the problem below. Ensure that your new answer completely avoids the same kinds of errors that appear in the examples.

% The problem is:
% \{question\}

% '''
% \end{tcolorbox}

% \section{Case Study for Memory Buffer}
% \label{appendix : memory}
% This section presents a case study demonstrating how the memory buffer reuses past flawed reasoning traces as hard negatives, guiding the model to avoid recurring errors and strengthen its reasoning robustness.
% \section{Error Patterns in LVLMs' Reasoning Processes}
% \begin{figure*}[t!]
%     \centering\includegraphics[width=\linewidth]{fig/error_pattern.pdf}
%     \caption{An example of flawed reasoning in parallel-line geometry, where incorrect angle correspondence leads to the wrong justification despite selecting the correct answer.}
%     \label{fig: err_1}
% \end{figure*}
% \begin{figure*}[t!]
%     \centering\includegraphics[width=\linewidth]{fig/error_Pattern_2.pdf}
%     \caption{A case of flawed reasoning in circle geometry, where misinterpreting arc ratios produces an incorrect justification though the final answer matches.}
%     \label{fig: err_2}
% \end{figure*}
% We provide further examples that illustrate the error patterns observed in the reasoning process during GRPO training. Fig. \ref{fig: err_1} highlights two typical issues: the Logical Soundness and  Correct Reasoning patterns. Fig. \ref{fig: err_2} highlights three typical issues: the Error Identification, Language Consistenc and Redundancy patterns.
\section{Reasoning Examples}
This section provides representative reasoning examples generated by PSO-7B, illustrating how the framework improves logical soundness and coherence while reducing systematic errors compared to baseline models.
\label{appendix : example}

\begin{figure*}[t]
    \centering
    % --- 左图 ---
    \begin{minipage}[t]{0.49\linewidth}
        \centering
        \includegraphics[width=\linewidth]{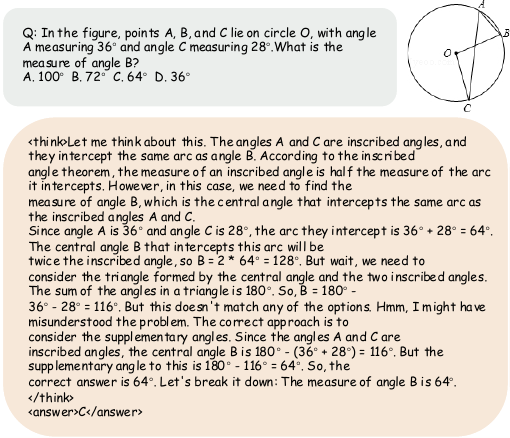}
        \caption{An PSO-7B reasoning example in circle geometry, showing step-by-step correction that leads to the correct angle measure.}
        \label{fig:demo_1}
    \end{minipage}
    \hfill
    % --- 右图 ---
    \begin{minipage}[t]{0.49\linewidth}
        \centering
        \includegraphics[width=\linewidth]{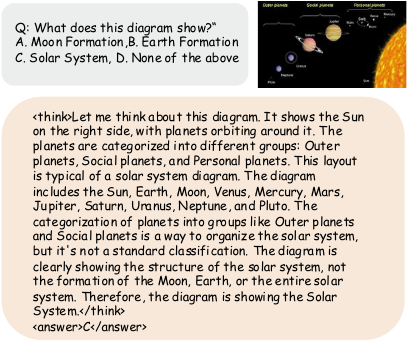}
        \caption{An PSO-7B reasoning example   a planetary diagram as the Solar System based on structure and categorization.}
        \label{fig:demo_22}
    \end{minipage}
\end{figure*}

\begin{figure*}[t]
    \centering
    % --- 左图 ---
    \begin{minipage}[t]{0.49\linewidth}
        \centering
        \includegraphics[width=\linewidth]{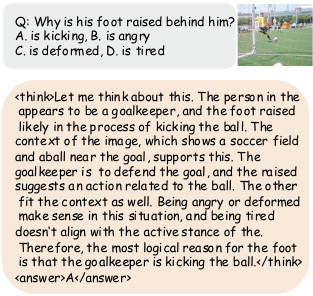}
        \caption{An PSO-7B multimodal reasoning example, correctly inferring from visual context that the goalkeeper’s raised foot indicates kicking the ball.}
        \label{fig:demo_11}
    \end{minipage}
    \hfill
    % --- 右图 ---
    \begin{minipage}[t]{0.49\linewidth}
        \centering
        \includegraphics[width=\linewidth]{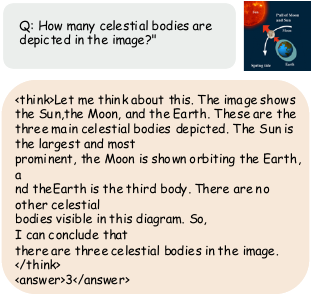}
        \caption{An PSO-7B reasoning example correctly identifying three celestial bodies - Sun, Moon, and Earth - in the diagram..}
        \label{fig:demo_2}
    \end{minipage}
\end{figure*}

% \begin{figure*}[t]
%     \centering
%     \includegraphics[width=0.9\linewidth]{fig/case_3.pdf}
%     \caption{A PSO-7B reasoning example on a planetary diagram, interpreting it as the Solar System based on structure and categorization.}
%     \label{fig:demo_3}
% \end{figure*}
% \begin{figure*}[t]
%     \centering
%     \includegraphics[width=0.9\linewidth]{fig/case_4.pdf}
%     \caption{An PSO-7B reasoning example correctly identifying three celestial bodies - Sun, Moon, and Earth - in the diagram.}
%     \label{fig:demo_4}
% \end{figure*}
% \section{Case Study on Sampling With and Without Memory-Integrated}
% \section{Case Study on Memory Experience Storage}

\section{Limitations and Future Work}
Although PSO substantially improves both answer accuracy and reasoning stability across diverse multimodal benchmarks, several limitations remain. First, the current thinking reward relies on self-evaluation from the base model, which may introduce bias or reward hacking behaviors when the model overfits its own scoring heuristics. Developing more robust and cross-model-consistent process reward models, especially those capable of multimodal introspection, is a promising direction. Second, the Negative Replay Memory stores only a small subset of low-reward trajectories, and the fixed-capacity FIFO strategy may overlook long-range or rare-but-critical reasoning failures. More expressive memory mechanisms—such as priority-based replay, structured error clustering, or task-aware memory routing—could further enhance stability.
Third, our online preference optimization operates at the path level but still treats each trajectory independently. Joint modeling of sets of reasoning paths or learning a global structure over the full trajectory distribution may unlock deeper insights into path selection bias. Additionally, PSO is evaluated primarily on vision-language reasoning tasks; its performance on long-horizon planning, multi-image reasoning, or interactive agent settings remains unexplored. Finally, the computational overhead of sampling multiple trajectories and performing self-critique per query is non-trivial. Designing more efficient trajectory sampling strategies, sparse reward mechanisms, or distillation methods to compress reasoning behaviors into lighter models will be important for real-world deployment.

Overall, we view PSO as an initial step toward reasoning-level alignment. Future work that strengthens reward robustness, expands memory-based error correction, and broadens applicability to richer multimodal and interactive environments could further advance the development of reliable and interpretable LVLM reasoning systems.

% \newpage
% \appendix
% \input{sec/7_appendix}

\end{document}